\documentclass[runningheads]{llncs}

\usepackage{eccv}

\usepackage{eccvabbrv}

\usepackage{graphicx}
\usepackage{booktabs}
\usepackage{tcolorbox}
\usepackage{multirow}
\usepackage{wrapfig}

\usepackage{colortbl}

\usepackage{makecell}
\usepackage{bbding}
\usepackage{pifont}
\usepackage{arydshln}
\usepackage{bbm}

\usepackage[accsupp]{axessibility}  
\usepackage[table]{xcolor} 

\usepackage{hyperref}

\usepackage{orcidlink}

\begin{document}

\title{The Fabrication of  Reality and Fantasy: Scene Generation with LLM-Assisted Prompt Interpretation} 

\titlerunning{Scene Generation with LLM-Assisted Prompt Interpretation}

\author{Yi Yao\inst{1*}\orcidlink{0000-0001-8227-5662} \and
Chan-Feng Hsu\inst{1*} \and
Jhe-Hao Lin\inst{1} \and
Hongxia Xie\inst{2}\orcidlink{0000-0002-5652-4327} \and
Terence Lin\inst{1} \and
Yi-Ning Huang\inst{1} \and
Hong-Han Shuai\inst{1} \orcidlink{0000-0003-2216-077X}\and
Wen-Huang Cheng\inst{3} \orcidlink{0000-0002-4662-7875}
}

\authorrunning{Y. Yao et al.}

\institute{National Yang Ming Chiao Tung University, Taiwan, \\
\email{\{leo81005.ee10, cfhsu311510211.ee11, hhshuai\}@nycu.edu.tw}
\and Jilin University, China, \\
\and National Taiwan University, Taiwan \\
\email{\{wenhuang@ntu.edu.tw\}}
}
\def\thefootnote{*}\footnotetext{These authors contributed equally to this work}\def\thefootnote{\arabic{footnote}}
\maketitle

\begin{abstract}

In spite of recent advancements in text-to-image generation, limitations persist in handling complex and imaginative prompts due to the restricted diversity and complexity of training data. This work explores how diffusion models can generate images from prompts requiring artistic creativity or specialized knowledge. We introduce the Realistic-Fantasy Benchmark (RFBench), a novel evaluation framework blending realistic and fantastical scenarios. To address these challenges, we propose the Realistic-Fantasy Network (RFNet), a training-free approach integrating diffusion models with LLMs. Extensive human evaluations and GPT-based compositional assessments demonstrate our approach's superiority over state-of-the-art methods. Our code and dataset is available at \url{https://leo81005.github.io/Reality-and-Fantasy/}.
\end{abstract}

\begin{figure*}
\vspace{-1em}
  \centering
  \centerline{\includegraphics[width=0.9\textwidth]{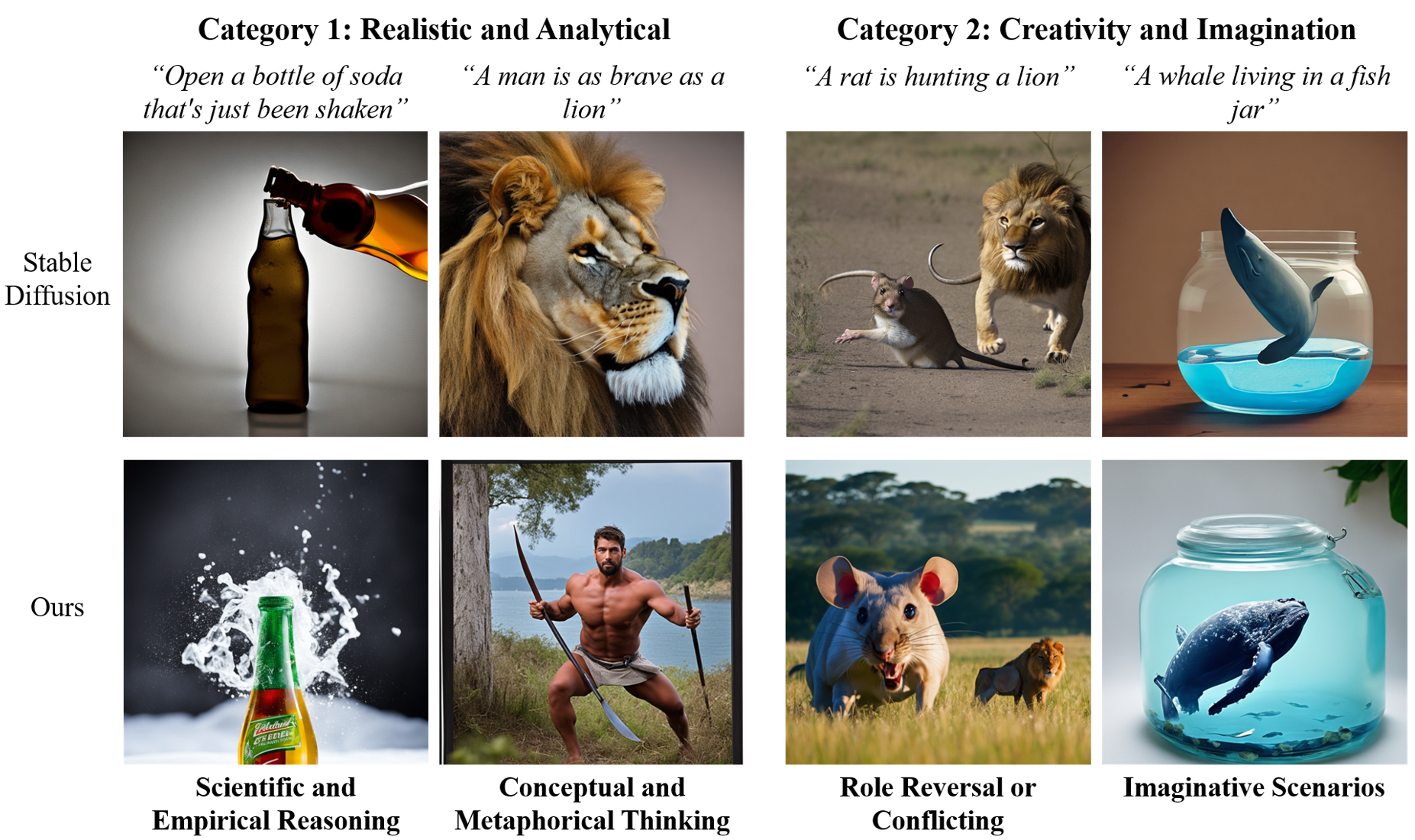}}
  \caption{Text-to-image diffusion models such as Stable Diffusion~\cite{rombach2022high} often struggle to accurately follow prompts that involve scientific and empirical reasoning, metaphorical thinking, role conflicting, or imaginative scenarios. Our method achieves enhanced prompt understanding capabilities and accurately follows these types of prompts.}
  \label{fig:fig1}
\vspace{-2em}
\end{figure*}

\section{Introduction}
\label{sec:intro}

Considerable advancements have been made in the field of text-to-image generation, especially with the introduction of diffusion models, \textit{e.g.}, Stable Diffusion~\cite{rombach2022high}, GLIDE~\cite{nichol2021glide}, DALLE2~\cite{ramesh2022hierarchical} and Imagen~\cite{saharia2022photorealistic}.
These models exhibit remarkable proficiency in generating diverse and high-fidelity images based on natural language prompts. 

However, despite their impressive capabilities, diffusion models occasionally face challenges in accurately interpreting complex prompts that demand a deep understanding or specialized knowledge~\cite{zhang2023survey, yang2024diffusion, zhang2023texttoimage}. 
This limitation becomes particularly apparent with creative and abstract prompts, which require a nuanced grasp of context and subtleties.
For example, in scenarios where the prompt involves unconventional scenarios like ``a rat is hunting a lion'', traditional diffusion models might not accurately represent the intended dynamics or relationships between entities (as shown in Fig.\ref{fig:fig1}).


A significant obstacle for traditional diffusion models in creating abstract images is the bias present within their training datasets~\cite{perera2023analyzing, friedrich2023fair}. These datasets often do not include images of scenarios that defy conventional reality, such as a mouse hunting a lion. 
Traditionally, mitigating these challenges has required costly data collection and complex filtering, as well as model retraining or fine-tuning~\cite{golnari2023lora, hu2021lora, zhang2023adding, yang2023reco}. Research costs significantly increase due to these long and labor-intensive processes. Moreover, fine-tuning neural networks and model editing can lead to catastrophic forgetting and overall performance degradation~\cite{kemker2018measuring, smith2023continual}. Recently, it has been demonstrated that utilizing Large Language Models (LLMs) to aid in the generation process ensures the production of accurate details~\cite{wu2023self, lian2023llmgroundedvideo, lian2023llmgrounded, yang2023reco, feng2024layoutgpt, yang2024mastering, qin2024diffusiongpt}. Directly integrating these models during the generation phase signifies a more efficient strategy.

In this paper, we want to address the question: \textit{how can generative models be improved to better capture imaginative and abstract concepts in images?} 
In response to the existing gap in benchmarks for abstract and creative text-to-image synthesis, our work introduces a novel benchmark, \textbf{Realistic-Fantasy Benchmark (RFBench)}. This benchmark is designed to evaluate both \textit{Realistic} \& \textit{Analytical} and \textit{Creativity} \& \textit{Imagination} interpretations in generated images. The \textit{Realistic} \& \textit{Analytical} category includes four sub-categories, focusing on the models' ability to adhere to realism and analytical depth. Images are generated in response to prompts that require not only precision in science but also cultural sensitivity and nuanced expression of symbolic meaning. On the other hand, \textit{Creativity} \& \textit{Imagination}, is segmented into five specific sub-categories based on attribute distinctions, challenges models to navigate the complexities of generating images from prompts that necessitate a high degree of creativity and abstract reasoning.

To empower diffusion models with the capability to generate imaginative and abstract images, we introduce an innovative training-free approach \textbf{Realistic-Fantasy Network (RFNet) }that integrates diffusion models with LLMs. 
Given a prompt describing the desired image, the LLM generates an image layout, which includes bounding boxes for main subjects and background elements, along with textual details to support logic or interpret scientific data.
To refine image generation, we further propose the Semantic Alignment Assessment (SAA), ensuring consistency with the scene's objects. This crucial step improves the final image quality. The enhanced details direct the diffusion model, enabling precise object placement through guidance constraints. Our method, leveraging pre-trained models, is compatible with independently trained LLMs and diffusion models, eliminating the need for parameter adjustments.
In summary, our key contributions are:
\begin{enumerate}
    \item We have collected a novel \textbf{Realistic-Fantasy Benchmark (RFBench)}, which is a meticulously curated benchmark that stands out for its rich diversity of scenarios. It challenges and extends the boundaries of generative model creativity and inference capabilities, establishing a new standard for assessing imaginative data processing.
    \item To empower diffusion models with the capability to generate imaginative and abstract images, we introduce an innovative training-free approach, \textbf{Realistic-Fantasy Network (RFNet)}, that integrates diffusion models with LLMs.
    \item Through our proposed RFBench, extensive human evaluations coupled with GPT-based compositional assessments have demonstrated our approach's superiority over other state-of-the-art methods. 
\end{enumerate}

\section{Related Work}

\subsection{Text-guided diffusion models}

Diffusion models, utilizing stochastic differential equations, have emerged as effective tools for generating realistic images \cite{rombach2022high, anciukevivcius2023renderdiffusion, gong2023diffpose, feng2023trainingfree, nair2023steered, ramesh2022hierarchical, nichol2021glide, saharia2022photorealistic}. 
DALL-E 2 \cite{ramesh2022hierarchical} pioneered the approach of converting textual descriptions into joint image-text embeddings with the aid of CLIP \cite{radford2021learning}. GLIDE \cite{nichol2021glide} demonstrated that classifier-free guidance \cite{ho2022classifier} is favored by human evaluators over CLIP guidance for generating images based on text descriptions.  Imagen \cite{saharia2022photorealistic} follows GLIDE but uses pretrained text encoder instead, further reducing negligible computation burden to the online training of the text-to-image diffusion prior, and can improve sample quality significantly by simply scaling the text encoder. 

Although text-to-image capabilities have seen significant development, there has been limited focus on generating images involving \textit{high levels of creativity, scientific principles, cultural references, and symbolic meanings.} The primary reason is the data bias in the training dataset~\cite{naik2023social, su2023unbiased, 10377171}.
Several studies~\cite{perera2023analyzing, luccioni2024stable,naik2023social} have investigated the impact of data bias on diffusion models, particularly in the context of Text-to-Image generation. Perera \textit{et al.}\cite{perera2023analyzing} investigates the bias exhibited by diffusion models across various attributes in face generation. 
Luccioni \textit{et al.}\cite{luccioni2024stable} evaluates bias levels in text-to-image systems regarding gender and ethnicity. 

In this work, we introduce a new task: reality and fantasy scene generation. Recognizing the absence of a dedicated evaluation framework for such tasks, we introduce a new benchmark, the Realistic-Fantasy Benchmark (RFBench), which blends scenarios from both realistic and fantastical realms.

\subsection{LLMs for image generation}

Recently, researchers have explored using LLMs to provide guidance or auxiliary information for text-to-image generation systems~\cite{qu2023layoutllm, mantri2023interactive, lian2023llmgrounded, gani2024llm, yang2024mastering, wu2023self, qin2024diffusiongpt}. 
In LMD \cite{lian2023llmgrounded}, foreground objects are identified using LLMs, and then images are generated based on the layout determined by the diffusion model. Phung \textit{et al. }\cite{phung2023grounded} proposes attention-refocusing losses to constrain the generated objects on their assigned boxes generated by LLMs. LVD \cite{lian2023llmgroundedvideo} requires LLMs to generate continuous spatial constraints to accomplish video generation.
Besides using LLMs to generate spatial layout from user prompts, some studies \cite{wu2023self, yang2024mastering, qin2024diffusiongpt} investigate integrating LLMs directly into the image generation pipeline. 
SLD \cite{wu2023self} integrates open-vocabulary object detection with LLMs to enhance image editing. 
RPG \cite{yang2024mastering} integrates LLMs in a closed-loop manner, allowing generated images to continuously improve through LLMs feedback, and uses Chain-of-Thought~\cite{wei2022chain} to further improve generation quality.

As a result of these developments, LLMs can be incorporated into pipelines for the generation of images.
In this work, we use LLMs to uncover and elaborate upon the complexities embedded within complex and abstract prompts. 

\section{Our Proposed Realistic-Fantasy Benchmark}
In this study, we explore how diffusion models can effectively process and generate imagery from prompts that pose significant challenges due to their reliance on creative thinking or specialized knowledge. Recognizing the absence of a dedicated evaluation framework for such tasks, we introduce a new benchmark, the \textbf{Realistic-Fantasy Benchmark (RFBench)}, which blends scenarios from both realistic and fantastical realms.  

\begin{figure*}
\vspace{-1em}
  \centering
\centerline{\includegraphics[width=0.8\textwidth]{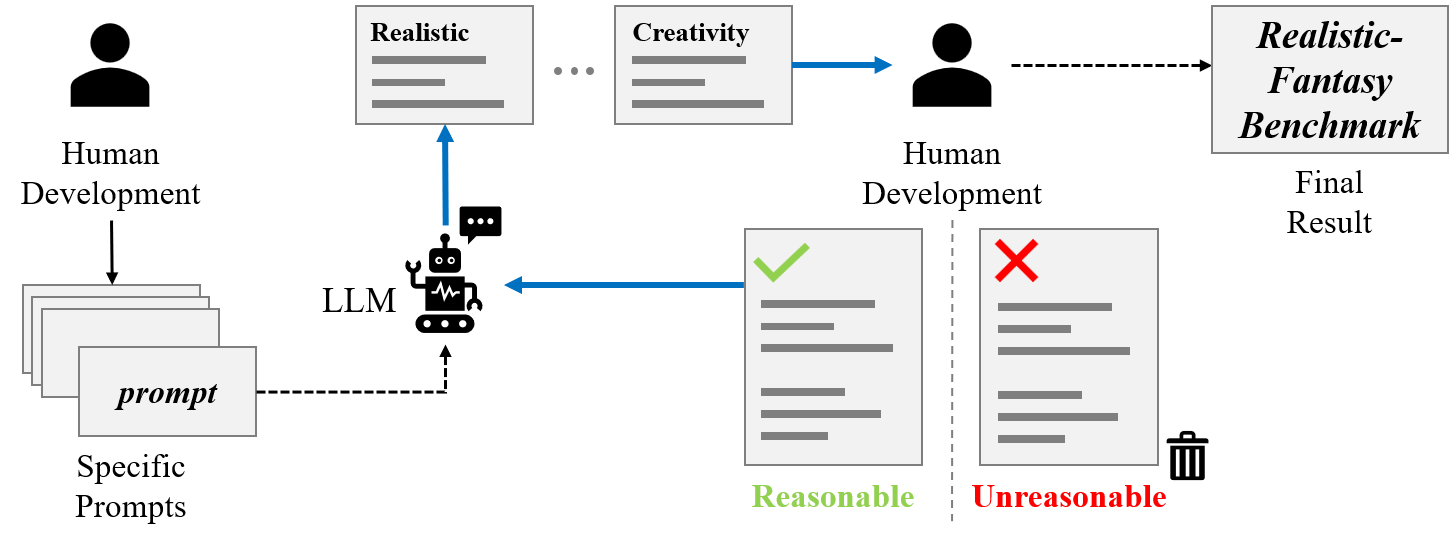}}
  \caption{The collection pipeline of our proposed \textbf{RFBench}.}
  \label{fig:benchmark}
\vspace{-2em}
\end{figure*}

\textbf{Benchmark Collection.} We focus on two main categories, each with distinct subcategories, \textbf{Realistic} \& \textbf{Analytical} and \textbf{Creativity} \& \textbf{Imagination}, totaling nine subcategories.
Each sub-category is meticulously crafted with around 25 text prompts, leading to an aggregate of 229 unique compositional text prompts designed to test the models against both conventional and unprecedented creative challenges. 
The collection process, outlined in \cref{fig:benchmark}, employs a hybrid method combining in-context learning and predefined rules, leveraging powerful language models such as ChatGPT and Bard for diverse text prompts creation. By alternating between these models, we achieve a diverse set of responses, capitalizing on the distinct advantages of each LLM.
It boosts the variety and complexity of prompts while reducing the reliance on manual labeling.


\begin{table*}[t]
\centering
\small
\caption{Categories of our proposed Realistic-Fantasy Benchmark (RFBench). The full dataset are show in our supplementary material.}
\label{tab:benchmark_category}
\resizebox{\textwidth}{!}{%
\begin{tabular}{@{} p{4cm} : p{7cm} : p{4.6cm} @{}}
\hline
\hline
\multicolumn{1}{c}{\textbf{Categories}} &
  \multicolumn{1}{c}{\textbf{Definition}} &
  \multicolumn{1}{c}{\textbf{Example}} \\ \hline
\multicolumn{3}{c}{\textbf{Realistic} \& \textbf{Analytical}} \\
\multicolumn{3}{c}{To evaluate models' comprehensive understanding of complex ideas, factual accuracy, and cultural insights} \\ \hline
Scientific and Empirical Reasoning &
  Relates to hypothesis testing, deduction, and scientific methodology &
  \textit{``A drop of water on the International Space Station.''} \\ \hline
Cultural and Temporal Awareness &
  Necessitates understanding and knowledge of particular cultural or historical events &
  \textit{``Children in costumes going door-to-door on October 31st.''} \\ \hline
Factual or Literal Descriptions &
  Evaluates the models' capacity to generate images that adhere closely to factual accuracy and realistic depiction &
 \textit{``A tank that's been sitting on the beach for 50 years.''} \\ \hline
Conceptual and Metaphorical Thinking &
  Focuses on the models' ability to comprehend and depict the underlying symbolic messages within the prompts &
  \textit{``A man is as brave as a lion.''} \\ 
  \hline
  \hline
\multicolumn{3}{c}{\textbf{Creativity} \& \textbf{Imagination}} \\
\multicolumn{3}{c}{To evaluate model's capability in employing creativity and abstract thinking} \\ \hline
Common Objects in Unusual Contexts &
  Challenges models' capacity to maintain object integrity while adapting to surreal environments &
  \textit{``A rubber duck sailing across a field of hot lava.''} \\ \hline
Imaginative Scenarios &
  Evaluate the models' abilities to craft scenes involving animals or humans in fantastical or unlikely scenarios &
  \textit{``An octopus playing chess with a seahorse.''} \\ \hline
Counterfactual Scenarios &
  Focuses on scenarios that defy conventional expectations of reality &
  \textit{``Fish swimming in the clouds.''} \\ \hline
Role Reversal or Conflicting &
  Stimulates consideration of perspectives and scenarios where typical roles or expectations are inverted &
  \textit{``A cat is chased by a mouse.''} \\ \hline
Anthropomorphic Scenarios &
  Examines the model's ability to imbue inanimate objects or phenomena with human-like characteristics &
  \textit{``A snowman building a friend in the blizzard.''} \\ \hline
\end{tabular}
}
\vspace{-2em}
\end{table*}

\textbf{Realistic} \& \textbf{Analytical} \textbf{Category}. There are four sub-categories: \textit{Scientific and Empirical Reasoning}, \textit{Cultural and Temporal Awareness}, \textit{Factual or Literal Descriptions}, and  \textit{Conceptual and Metaphorical Thinking} (details are shown in the upper part of \cref{tab:benchmark_category}). 
These sub-categories are anchored in real-world contexts, emphasizing logical reasoning, accurate data, and an understanding of cultural or historical contexts. They contain scientific exploration,
realistic descriptions, and culturally symbolic narratives. 
This demands that the models not only draw from an extensive knowledge pool but also demonstrate an ability to grasp and articulate underlying concepts.

\textbf{Creativity} \& \textbf{Imagination} \textbf{Category}.
It consists of five sub-categories: \textit{Common Objects in Unusual Contexts}, \textit{Imaginative Scenarios}, \textit{Counterfactual Scenarios}, \textit{Role Reversal or Conflicting}, and \textit{Anthropomorphic Scenarios} (details are shown in the lower part of \cref{tab:benchmark_category}).  
This evaluation focuses on the model's capacity to innovatively repurpose familiar objects, attribute human-like characteristics to inanimate objects, and generate novel environments for everyday items. 
This category tests the model's out-of-the-box thinking and imaginative capabilities.

\section{Our Proposed Realistic-Fantasy Network}
\begin{figure*}[t]
  \centering
  \centerline{\includegraphics[width=0.93\textwidth]{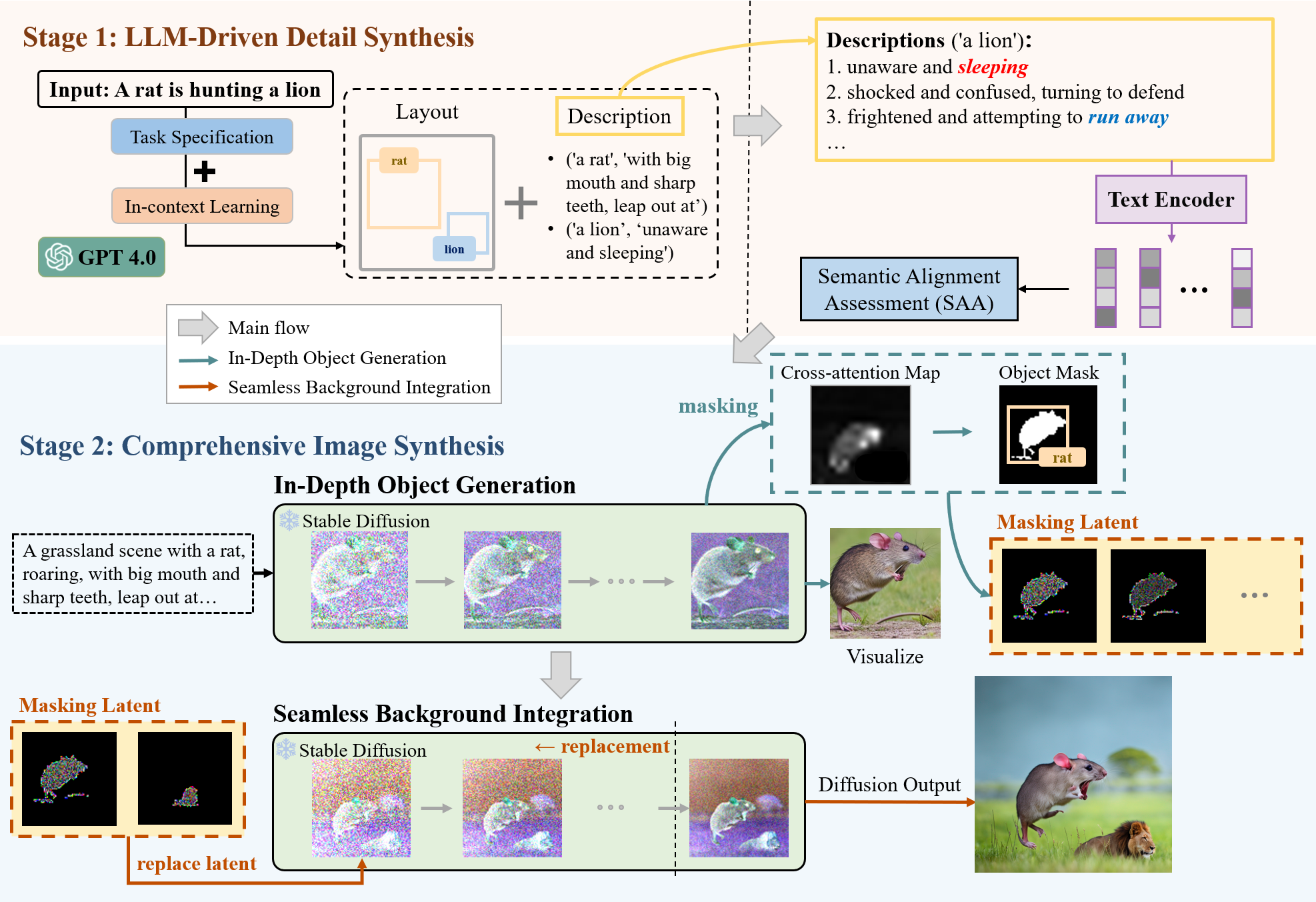}}
  \caption{\textbf{Overview of our proposed Realistic-Fantasy Network (RFNet).} In stage 1, the user's input prompt is first processed by a LLM to extract the layout and descriptions. The descriptions then go through a text encoder, which is the text-processing component of the CLIP model, and are refined by the SAA to form a better prompt. In stage 2, the refined prompts are fed into the diffusion model for in-depth object generation, which creates each target object with precision. The resulting cross-attention map and mask latent are then utilized for seamless background integration, merging objects into one single image.}
  \label{fig:framework}
\vspace{-0.5cm}
\end{figure*}

In this section, we propose a \textbf{Realistic-Fantasy Network (RFNet)} for the benchmark scenario we proposed in the previous section.
To thoroughly interpret the details from the input prompt, we divide our approach into two stages, as shown in \cref{fig:framework}. In the first stage, we transform the initial input prompt into a refined version specifically tailored for image generation by LLMs. In the second stage, we utilize a diffusion model through a two-step process to generate outputs with extraordinary details.

\subsection{LLM-Driven Detail Synthesis}


In the first stage of our methodology, we concentrate on utilizing LLMs to uncover and elaborate on the intricacies embedded within the user's input prompt. 
This process involves specifying task requirements to more accurately define the task and incorporating in-context learning to enhance understanding for LLMs. The enriched response from the LLM encompasses additional information, such as \textit{layout, detailed descriptions, background scenes,} and \textit{negative prompts}~\footnote{One detailed sample can be found in our supplementary material.}. 
This step is crucial as it aims to mitigate the primary challenge we seek to overcome: the training data bias inherent in current diffusion models. By leveraging the pre-trained LLM for logical reasoning and conjecture, we aim to compensate for the gaps left by these biases, ensuring a more accurate and coherent image generation process.

\subsection{Semantic Alignment Assessment}


As we proceed with generating images using the diffusion model using the details generated by the previous step, there is a critical challenge: \textit{the description lists generated by LLMs for one object usually overlook the relationships among them.} For example, interpretations of ``a lion'' could range from being ``unaware and asleep'' to ``frightened and trying to escape.'' Although both depictions are valid, descriptions such as ``unaware'' and ``trying to escape'' can lead to conflicting interpretations, thus complicating the image generation process.

To overcome this challenge, we introduce the \textbf{Semantic Alignment Assessment (SAA)} module. This module calculates the relevance between different object vectors, thereby selecting the candidate description that best fits the current scenario. 
By conducting the cosine similarity among different descriptions, we can navigate the complexities introduced by the LLM's output, selecting the most compatible details for the diffusion model. This step is crucial for maintaining the coherence and accuracy of the generated images, highlighting our novel approach to mitigating the risk of conflicting descriptions. 
Through this module, we ensure textual precision and compatibility, and provide \textit{clear, consistent instructions} for the subsequent diffusion model to generate visually coherent representations.



\begin{figure}
  \centering
  \centerline{\includegraphics[width=1\textwidth]{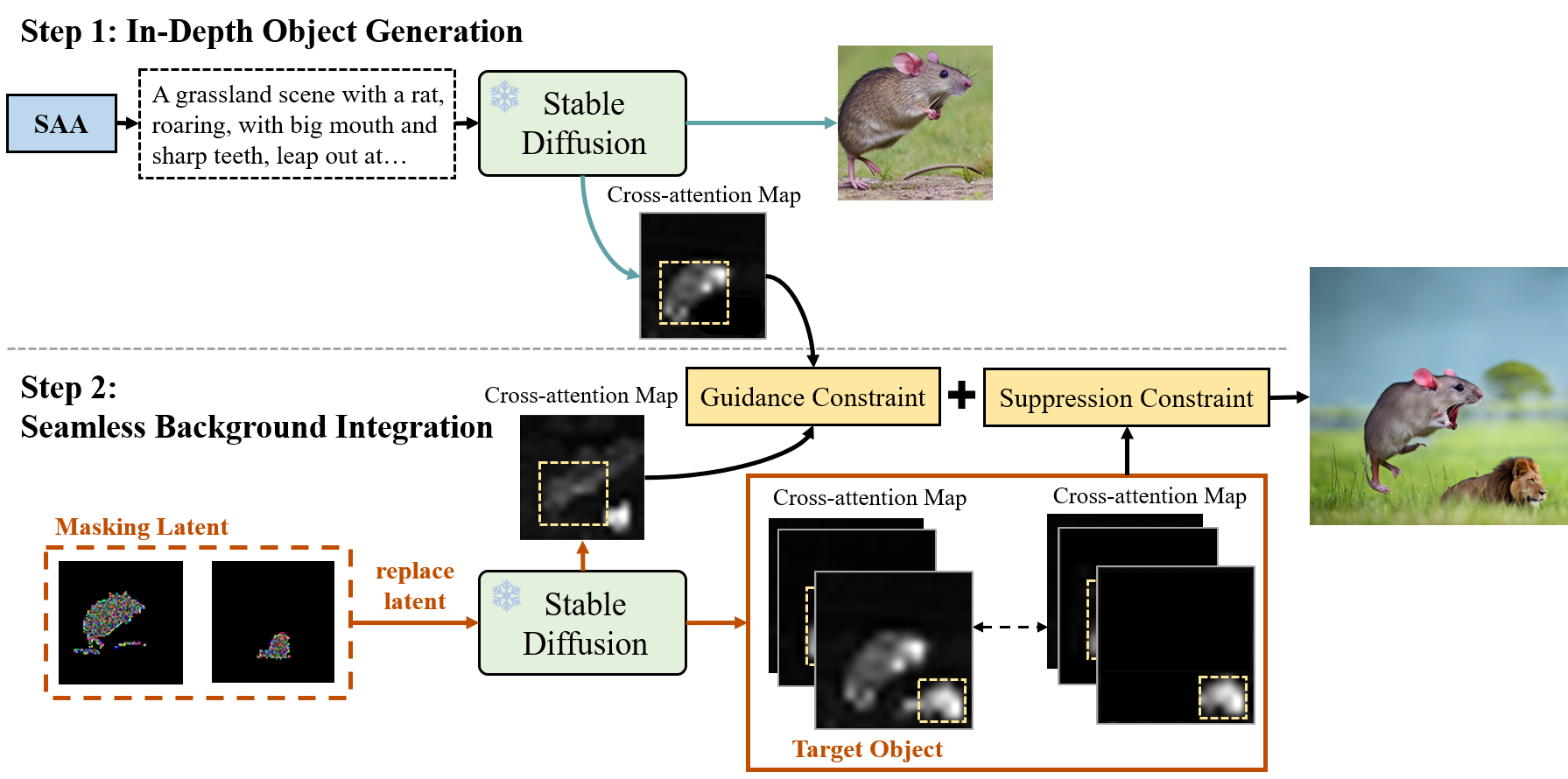}}
  \caption{\textbf{Comprehensive Image Synthesis}. In step 1, utilizing the prompt refined by the SAA module, the frozen stable diffusion model generates each foreground object independently. During the denoising phase, the cross-attention map is extracted and saved for \textbf{Guidance Constraint }in the next step. In addition, a \textbf{Suppression Constraint} is also added in step 2 to minimize influence between different objects.}
  \label{fig:image_generation}
\vspace{-0.6cm}
\end{figure}

\subsection{Comprehensive Image Synthesis}
In the second stage of our proposed RFNet, following LMD~\cite{lian2023llmgrounded}, we propose a two-step generation process for imaginative and abstract concepts.
As shown in \cref{fig:image_generation}, in the first step, we focus on generating each \textit{foreground} object with comprehensive details. In the second step, we integrate the objects generated in the first step into corresponding \textit{background} derived from the initial prompt. This structured approach ensures a cohesive integration of detailed foreground objects into a contextually relevant background, enhancing the overall effectiveness of our framework.

\textbf{Step 1: In-Depth Object Generation.}

We re-organize the SAA description lists to consider both the layout of specific objects and their descriptions. By concatenating the background prompt with the target object and its relevant descriptions, we set up the input prompt as \textit{``[background prompt] with [target object], [descriptions]''} (e.g., ``A grassland scene with a rat, roaring, with big mouth and sharp teeth, leap out at...''). Following LMD \cite{lian2023llmgrounded}, the initial latent representation for each target object is fixed to facilitate the fusion of various objects into a cohesive background scene. 

During generation, the diffusion model uses cross-attention layers to manage the influence of textual information on the visual output, allowing precise control over image details. The cross-attention map's constraint function integrates objects within the bounding box by enhancing cross-attention inside the box for accurate object representation while minimizing it outside the box. This function guides the update of the noised latent vector during denoising to ensure spatial conditions match predefined specifications. The constraint function is defined as:

\begin{equation}
    \mathcal{L}_{obj}(\textbf{A}, i, v) = [1-\rm{Topk}_u(\textbf{A}_{uv} \cdot \textbf{m}_i)] + [\rm{Topk}_u(\textbf{A}_{uv} \cdot (1-\textbf{m}_i))],
\end{equation}

\noindent where \(\textbf{m}_i\) denotes the binary mask of the bounding box associated with object \(i\), performing element-wise multiplication over the cross-attention map \(\textbf{A}\). The cross-attention map \(\textbf{A}\) is aggregated by summing the contributions across all layers. For object \(i\), the operation \(\rm{Topk}_u\) computes the mean of the top-k values within the spatial dimension \(u\). Prior to each denoising step, the latent is refined by minimizing the constraint function:

\begin{equation}
    z_t^{'} \leftarrow z_t - \alpha \cdot \nabla_{z_t} \sum_{v \in V} \mathcal{L}(\textbf{A}, i, v),
\end{equation}
\begin{equation}\label{eq:3}
    z_{t-1} \leftarrow Diffusion Step(z_t^{'}, \mathcal{P}^{(i)}),
\end{equation}

\noindent where \(\alpha\) denotes the hyperparameter that controls the magnitude of the gradient update, and \(V\) contains the set of token indices for the target object in the prompt. As in the diffusion step, the updated \(z_t^{'}\) along with the modified prompt \(\mathcal{P}^{(i)}\) of object \(i\), are served as the inputs to the diffusion model.

After the generation, the cross-attention map derived from each target object is then converted into a saliency mask. This mask is applied to the latent representation of the target object through element-wise multiplication at each step of the denoising process. Both the cross-attention map and the masked latent representation of the target object between each denoising step are transmitted to the next step for background integration.

\textbf{Step 2: Seamless Background Integration.} This step involves fusing the generated results with the background while preserving the high-quality generation achieved in the first step. To accomplish this, we first replace the generated latent \(z_t^{'}\) with the masking latent \(z_t^{(masked, i)}\) for each object \(i\):

\begin{equation}
    z_t^{'} \leftarrow Replacement(z_t^{'}, z_t^{(masked, i)}), \ \forall i.
\end{equation}

Following the approach established in Step 1, the initial latent representation \(z_T^{'}\) is also fixed with the initial latent of each target object in the first step. This alignment ensures a seamless integration of the object into the background. The replacement approach is performed within timestep \(rT\), where \(r\in[0,1]\), reserving principles from LMD \cite{lian2023llmgrounded}. According to LMD, the diffusion model determines the position of the objects in the early denoising steps, while adjusting details in the later steps. This helps us to preserve exceptional control over the layout.

Furthermore, we incorporate a specialized constraint function designed to enhance the integration of generated objects with their background, distinguished by two key components: \textbf{guidance constraint} and \textbf{suppression constraint}. As shown in \cref{fig:image_generation}, the {guidance constraint} is engineered to reduce the cross-attention within each bounding box relative to the original object's attention. With the purpose of seamlessly integrating with the detailed object generated in the first step. Conversely, the \textit{suppression constraint} works to minimize cross-attention outside the bounding box, thereby mitigating interference among multiple objects when processed together, as illustrated in \cref{eq:5}. These constraint functions mark a departure from conventional methods that predominantly use loss to fix the layout. 


\begin{equation}\label{eq:5}
    \mathcal{L}_{bg}(\textbf{A}^{'}, \textbf{A}^{(i)}, i, v) = \underbrace{\beta \cdot \sum_{u} \left| (\textbf{A}_{uv}^{'} - \textbf{A}_{uv}^{(i)}) \cdot \textbf{m}_i \right|}_{\text{guidance constraint}} + \underbrace{\gamma \cdot \mathrm{Topk}_u(\textbf{A}_{uv}^{'} \cdot (1-\textbf{m}_i))}_{\text{suppression constraint}}, \ \forall i,
\end{equation}

\noindent where \(\textbf{A}^{'}\) represents the cross-attention map post the substitution procedure and \(\textbf{A}^{(i)}\) is the cross-attention map of object \(i\) extracted from the diffusion step in \cref{eq:3}. The hyperparameters \(\beta\) and \(\gamma\) indicate the intensity of guidance constraint and suppression constraint, respectively. It is noteworthy that adjusting  \(\beta\) amplifies the importance of the guidance loss within the overall loss function. This adjustment ensures the latent representation is precisely aligned with the object generated in the initial phase, thereby guaranteeing a high level of accuracy and consistency in the integration process. 
Upon completion of all denoising steps, the latent \(z_0\) is fed into the decoder to produce the final image.

Our strategy focuses on maintaining the integrity and coherence of the foreground objects generated, emphasizing the preservation of their quality and interaction with the background. By doing so, the generated visual elements are fidelity-aware and contextually appropriate.

\section{Experiments}

\subsection{Implementation Details}

\textbf{Experimental Setup.} In this work, we choose versions 1.4 and 2.1 of Stable Diffusion~\cite{rombach2022high} as the text-to-image baseline model. The number of denoising steps is set as 50 with a fixed guidance scale of 7.5, and the synthetic images are in a resolution of 512 × 512. All experiments are conducted on the NVIDIA RTX 3090 GPU with 24 GB memory.

\noindent \textbf{Evaluation Metrics.} We generate 32 images for each text prompt in \textbf{RFBench} for automatic evaluation.
We selected the following two metrics:
(1) GPT4-CLIP~\footnote{We adopt GPT4-CLIP due to BLIP-CLIP's~\cite{chefer2023attend} limitations in accurately capturing image meanings through generated captions.}. By utilizing GPT4 for captioning and calculating CLIP text-text cosine similarity, GPT4-CLIP ensures a more precise reflection of the intended meanings between images and prompts.  
(2) GPT4Score. Inspired by ~\cite{huang2024t2i, liu2023llava}, we adopt GPT4Score to evaluate image alignment with text prompts, where GPT4 rates images on a 0-100 scale based on their fidelity to the prompts, enabling precise assessment of model-generated visuals against specified criteria\footnote{The widely recognized metric, CLIPScore~\cite{radford2021learning, hessel2021clipscore} exhibits limitations in evaluating our task. For detailed examples, please see the supplementary materials.}. 

\definecolor{graybg}{gray}{0.90}
\begin{table*}
\centering
\small
\caption{Benchmarking on GPT4-CLIP and GPT4Score. R \&\ A, C \&\ I, Avg represent Realistic \&\ Analytical category, Creativity \&\ Imagination category, and average of both categories, respectively. The red text indicates the improvement ratio of our method compared to Stable Diffusion.}
\resizebox{\textwidth}{!}{%
\begin{tabular}{c|ccc|ccc}
\toprule
\multirow{2}{*}{Model} & \multicolumn{3}{c|}{GPT4-CLIP} & \multicolumn{3}{c}{GPT4Score} \\
& R \&\ A & C \&\ I & Avg & R \&\ A & C \&\ I & Avg \\
\hline
Stable Diffusion~\cite{rombach2022high}       & 0.573 & 0.552 & 0.561 & 0.667 & 0.440 & 0.541 \\
MultiDiffusion~\cite{bar2023multidiffusion}         & 0.510 & 0.510 & 0.510 & 0.517 & 0.493 & 0.504 \\
Attend and Excite~\cite{chefer2023attend}      & 0.523 & 0.560 & 0.546 & 0.633 & 0.520 & 0.570 \\
LLM-groundedDiffusion~\cite{lian2023llmgrounded}  & 0.457 & 0.536 & 0.501 & 0.550 & 0.600 & 0.578 \\
BoxDiff~\cite{xie2023boxdiff}                & 0.532 & 0.553 & 0.543 & 0.583 & 0.520 & 0.548 \\
SDXL~\cite{podell2023sdxl}                   & 0.536 & 0.619 & 0.582 & 0.567 & 0.587 & 0.578 \\
\rowcolor{graybg} \textbf{RFNet(ours)}                   & \textbf{0.587} \textcolor{red}{(2\%$\uparrow$)} & \textbf{0.623} \textcolor{red}{(13\%$\uparrow$)} & \textbf{0.607} \textcolor{red}{(8\%$\uparrow$)} & \textbf{0.833} \textcolor{red}{(25\%$\uparrow$)} & \textbf{0.627}\textcolor{red}{(43\%$\uparrow$)} & \textbf{0.719} \textcolor{red}{(33\%$\uparrow$)} \\
\bottomrule
\end{tabular}
}
\vspace{-2em}
\label{tab:table2}
\end{table*}


\noindent\textbf{Comparison with Existing Methods.} 
We benchmark our proposed RFNet against various open-source scene generation methods, including Stable Diffusion~\cite{rombach2022high}, Attend and Excite~\cite{chefer2023attend}, LMD~\cite{lian2023llmgrounded}, BoxDiff~\cite{xie2023boxdiff}, MultiDiffusion~\cite{bar2023multidiffusion}, and SDXL~\cite{podell2023sdxl}. 
Notably, all methods, including ours, utilize Stable Diffusion 2.1 as the foundational model, ensuring a fair comparison.

\subsection{Quantitative Evaluation}
\vspace{-0.1cm}
\textbf{Evaluation on RFBench.}
As evidenced in~\cref{tab:table2}, our approach significantly outperforms other methods for both {Realistic }\& {Analytical} and {Creativity} \& {Imagination} tasks. For {Realistic }\& {Analytical} task, our method seamlessly integrates LLM-based insights, achieving a remarkable accuracy improvement. Unlike Attend-and-excite, which focuses on semantic guidance, our method ensures precise adherence to detailed and complex prompt requirements. For the {Creativity} \& {Imagination}, which demands high degrees of creativity and abstract conceptualization, our method outperforms others by not only adhering to the imaginative aspects of prompts but also maintaining coherent structure and contexts.
For instance, SDXL, while adept at high-resolution image synthesis, occasionally lacks in capturing the nuanced creativity intended in prompts; our method fills this gap effectively. Similarly, LMD, though enhancing prompt understanding through LLMs, sometimes struggles with the scientific reasoning required for {Realistic}\& {Analytical} tasks. 

Notably, for {Realistic }\& {Analytical} category, our approach shows a 61\% performance increase over MultiDiffusion on GPT4Score. Meanwhile, in {Creativity} \& {Imagination} task, we observe a substantial enhancement, outperforming Stable Diffusion by over 43\%. 
In light of the above, our method is unique in its ability to bridge the gap between realistic reasoning and imagination, creating a new benchmark for text-to-image generation.

\begin{wraptable}{r}{0.5\linewidth}
    \vspace{-3em}
    \caption{GPT4Score comparison with Imagen on DrawBench subset.}
    \label{tab:table3}
    \scriptsize
    \begin{tabular}{l|c|c}
    \toprule
    Prompt & Imagen & \textbf{Ours}\\
    \hline
    A bird scaring a scarecrow      & 0.069 & \textbf{0.275} \\ 
    A blue coloured pizza           & \textbf{0.425} & 0.125 \\
    A fish eating a pelican         & 0.000 & 0.000 \\
    A horse riding an astronaut     & 0.000 & 0.000 \\
    A panda making latte art        & 0.050 & \textbf{0.250} \\
    A pizza cooking an oven         & 0.700 & \textbf{0.831} \\
    A shark in the desert           & 0.194 & \textbf{0.713} \\
    An elephant under the sea       & 0.300 & \textbf{0.900} \\
    Hovering cow abducting aliens   & 0.025 & \textbf{0.144} \\
    Rainbow coloured penguin        & 0.394 & \textbf{0.519} \\
    \bottomrule
    \end{tabular}
    \vspace{-0.5cm}
\end{wraptable}
\noindent\textbf{Evaluation on DrawBench}. 
We also evaluate our method on DrawBench~\cite{saharia2022photorealistic}, a comprehensive and challenging benchmark for text-to-image models. Similar to us, DrawBench also includes some {Creativity} \& {Imagination} prompts, and we evaluate our method with Imagen~\cite{saharia2022photorealistic} on these prompts. As shown in~\cref{tab:table3}, our approach significantly outperforms Imagen on most prompt settings, demonstrating the generalization ability of our model.


\subsection{Qualitative Evaluation}

\begin{figure*}[t]
  \centering
  \centerline{\includegraphics[width=1.05\textwidth]{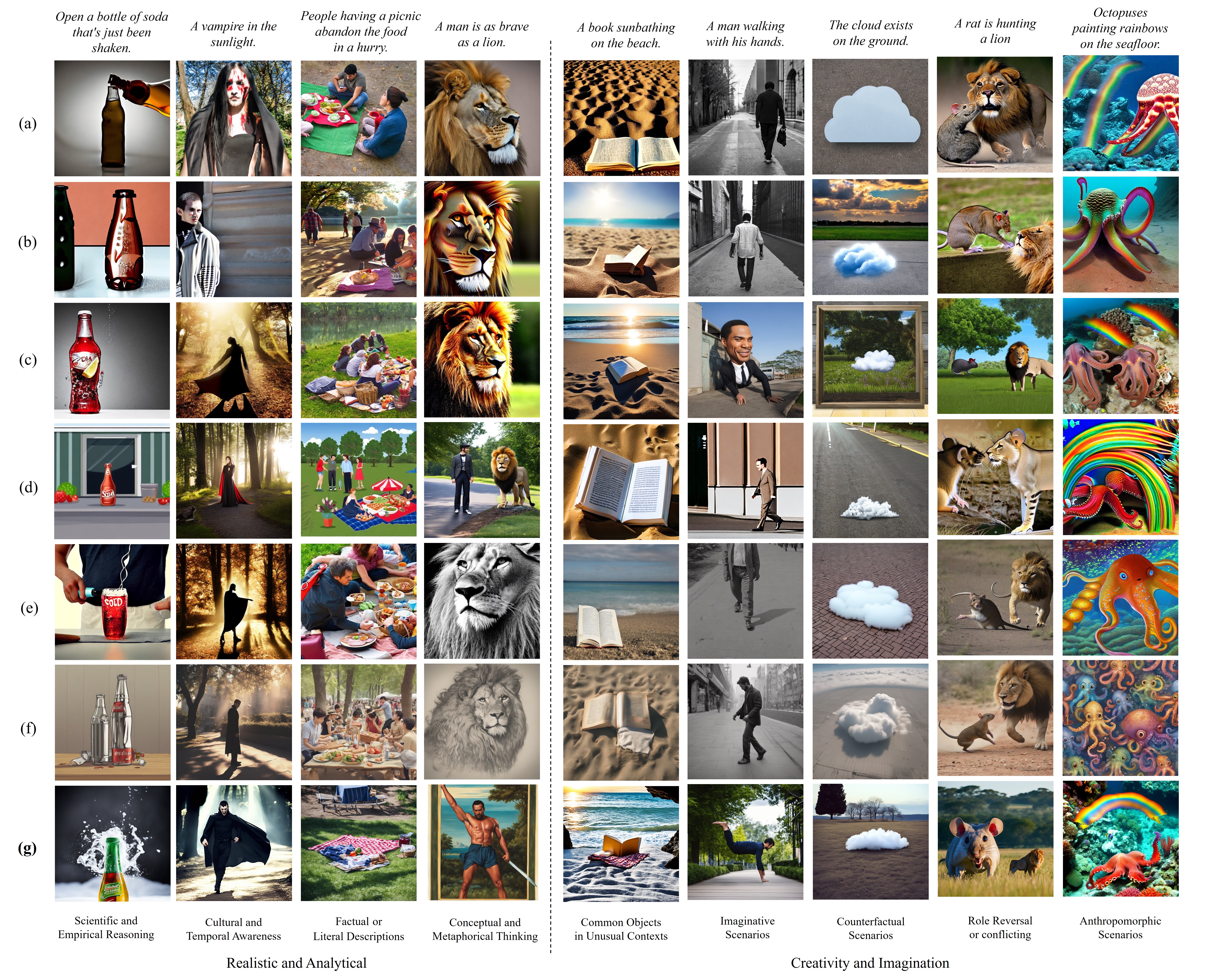}}
  \caption{Qualitative comparison on RFBench. The compared models include (a) Stable Diffusion, (b) MultiDiffusion, (c) Attend and Excite, (d) LMD, (e) BoxDiff, (f) SDXL, (g) Ours (Best viewed in color and zoom in. More samples can be found in our supplementary material.)}
  \label{fig:fig5}
\vspace{-1em}
\end{figure*}

In the qualitative comparison of text-guided image generation, we select some advanced baseline methods, including Attend-and-excite~\cite{chefer2023attend}, BoxDiff~\cite{xie2023boxdiff}, LMD~\cite{lian2023llmgrounded}, MultiDiffusion~\cite{bar2023multidiffusion}, and SDXL~\cite{podell2023sdxl}. 
Attend-and-excite focuses on enhancing the semantic understanding of prompts through attention mechanisms, while BoxDiff introduces a novel approach to text-to-image synthesis with box-constrained diffusion without the need for explicit training. MultiDiffusion proposes a method for fusing multiple diffusion paths to achieve greater control over the image generation process, and SDXL aims at improving the capabilities of latent diffusion models for synthesizing high-resolution images. As shown in Fig.~\ref{fig:fig5}, our method, produces more precise editing results than the aforementioned methods. This is attributed to our \textbf{In-Depth Object Generation} and \textbf{Seamless Background Integration} strategy. It ensures outstanding fidelity in outcomes and flawlessly retains the semantic structure of the source image, highlighting our approach's superior capability in complex editing tasks.

\subsection{User study}

Through an extensive user study, we benchmarked our model against other methods to assess real human preferences for the generated images. 
Utilizing our newly proposed benchmark, the RFBench, we selected a diverse set of 27 prompts and generated six images per prompt to ensure a broad representation of the model's capabilities. 
Detailed feedbacks were collected from 120 participants, evaluating each image for visual quality and text prompt fidelity~\footnote{Details of survey samples can be found in our supplementary material.}. These criteria are critical, which measure the image's quality and correctness of semantics in the synthesized image. 
Participants rated images on a scale from \{1, 2, 3, 4, 5\}, with scores normalized by dividing by 5. We calculated the average score across all images and participants.

\begin{figure}
\vspace{-1em}
    \centering
    \begin{minipage}{.3\textwidth}
        \begin{subfigure}{\textwidth}
        \centering
        \includegraphics[width=\textwidth]{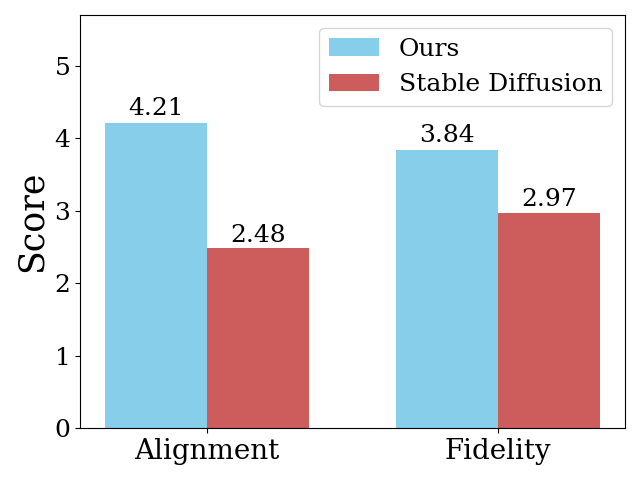}
        \end{subfigure}\\
        \begin{subfigure}{\textwidth}
        \centering
        \includegraphics[width=\textwidth]{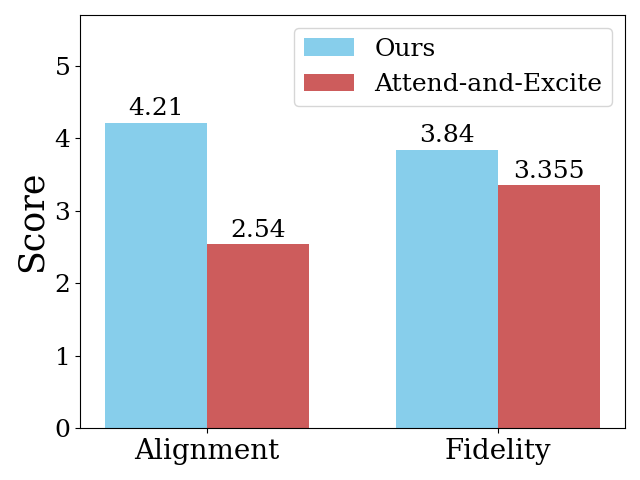}
        \end{subfigure}    
    \end{minipage}
    \hfill
    \begin{minipage}{.3\textwidth}
        \begin{subfigure}{\textwidth}
        \centering
        \includegraphics[width=\textwidth]{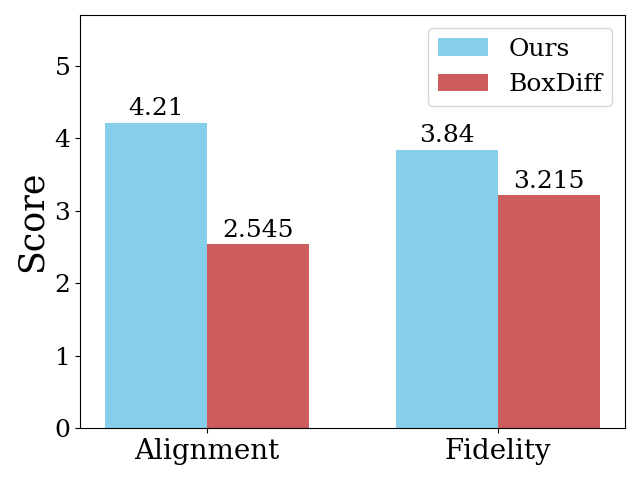}
        \end{subfigure}\\
        \begin{subfigure}{\textwidth}
        \centering
        \includegraphics[width=\textwidth]{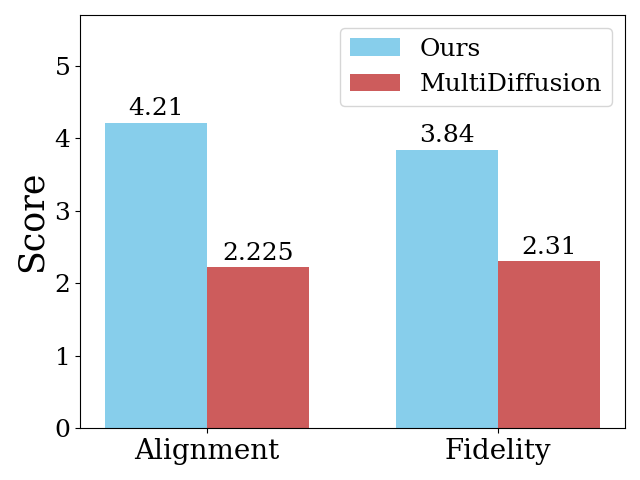}
        \end{subfigure}    
    \end{minipage}
    \hfill
    \begin{minipage}{.3\textwidth}
        \begin{subfigure}{\textwidth}
        \centering
        \includegraphics[width=\textwidth]{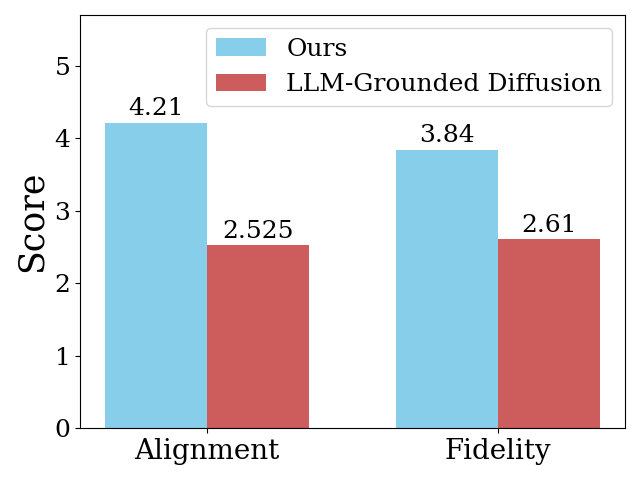}
        \end{subfigure}\\
        \begin{subfigure}{\textwidth}
        \centering
        \includegraphics[width=\textwidth]{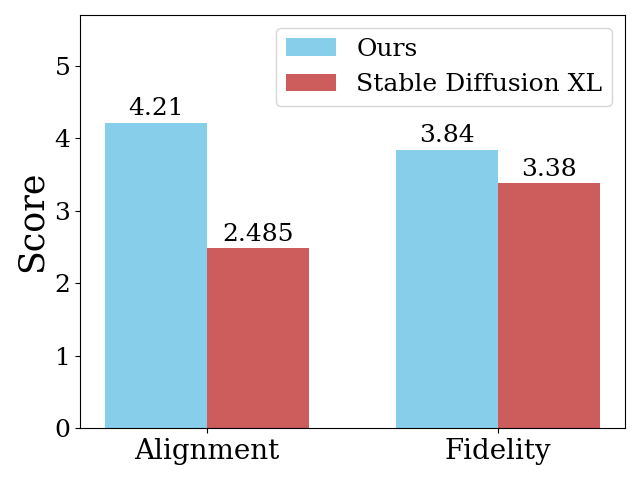}
        \end{subfigure}    
    \end{minipage}
    \caption{Comparison between our method and other advanced methods on RFBench. The image-text \textit{Alignment} and \textit{Fidelity} of our method are highly preferred by users.} 
\label{fig:fig6}
\vspace{-1.0cm}
\end{figure}

As illustrated in Fig.~\ref{fig:fig6}, 
participants uniformly favored our model's output, recognizing it as superior in both quality and alignment with the textual descriptions. 

\setlength\intextsep{3pt}
\setlength{\columnsep}{10pt}



\subsection{Ablation study}

\begin{wraptable}{r}{0.5\linewidth}
\vspace{-1cm}
\centering
\small
    \caption{Ablation studies on various components on RFBench.}
    \label{tab:table4}
    \begin{tabular}{c|c|c|c}
\hline
SAA & guidance & suppression & GPT4Score \\ \hline
             &              &              & 0.295 \\
             &             & $\checkmark$ & 0.407 \\
& $\checkmark$ &              & 0.554 \\
 & $\checkmark$ & $\checkmark$ & 0.572 \\
$\checkmark$ & $\checkmark$ & $\checkmark$ & \textbf{0.719}\\ \hline
\end{tabular}%
\end{wraptable}
\textbf{Impact of Various Constraints.}
To validate the impact of {guidance constraint} and {suppression constraint}, we perform ablation studies on different combinations of constraints, and the results are listed in~\cref{tab:table4}. As shown, the baseline model (Stable Diffusion) achieves a 0.295 in terms of GPT4Score without any constraints. As {guidance constraint} and {suppression constraint} work complementary to restrict the cross-attention of objects inside the conditional boxes, a higher GPT4Score of 0.572 is achieved on the generated images. Both proposed constraints are effective in controlling image quality and layout of synthesized foreground objects.

\begin{figure*}
\vspace{1.5em}
  \centering
  \centerline{\includegraphics[width=0.9\textwidth]{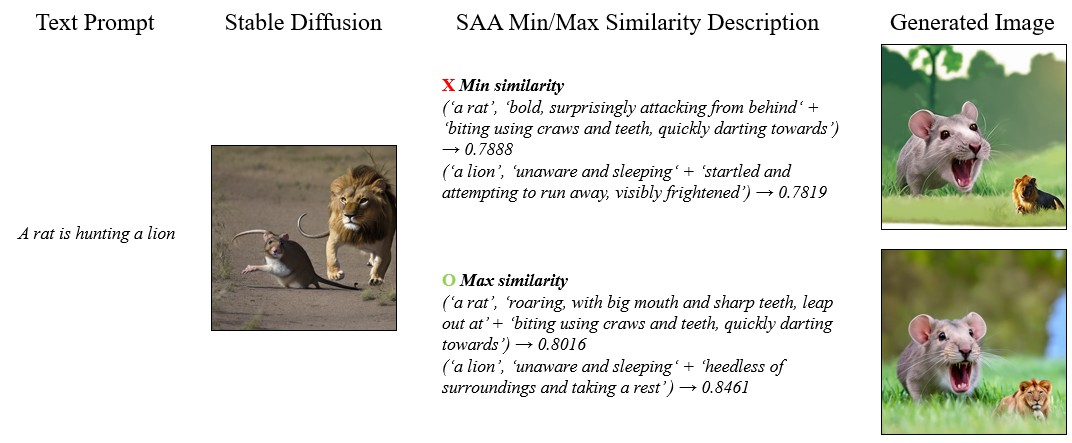}}
  \caption{The illustration of final generated results of minimum (min) and maximum (max) similarity descriptions during the SAA. It can be observed that prompts with higher similarity yield images of higher quality, which is significantly better than the one with the lowest similarity.}
  \label{fig:SAA}
\vspace{-1em}
\end{figure*}

\noindent \textbf{Impact of Semantic Alignment Assessment (SAA) Module.}
As aforementioned, using conflict descriptions in the denoise step may potentially affect image synthesis. The quantitative evaluation is presented in~\cref{tab:table4}. In the absence of SAA, the model attains a GPT4Score of 0.572. Similarly, with SAA, the model reaches a GPT4Score of 0.719. This indicates a lack of consistency between the semantics generated in the images and the provided text prompts, leading to a reduction in image quality. It is important to note that the inclusion of SAA significantly enhances the clarity of the images obtained. One visual illustration can be found in ~\cref{fig:SAA}.

\section{Conclusion and Future Work}
In this research, we present a novel challenge: generating scenes that blend reality and fantasy. We investigate the capacity of diffusion models to create visuals from prompts that demand high levels of creativity or specific knowledge. Noting the lack of a specific evaluation mechanism for such tasks, we establish the Realistic-Fantasy Benchmark (RFBench), combining elements of both realistic and imaginary scenarios. To address the task of generating realistic and fantastical scenes, we introduce a unique, training-free, two-tiered method, Realistic-Fantasy Network (RFNet), that combines diffusion models with large language models (LLMs).
Our approach, evaluated through the RFBench using thorough human assessments and GPT-based compositional evaluations, has proven to be superior to existing cutting-edge techniques.
Given the novelty of our task, future research could develop additional evaluation metrics beyond those used in this study, enhancing the assessment of generated scenes. 


%
%
\bibliographystyle{splncs04}
\bibliography{main}

\newpage
\renewcommand\thesection{\Alph{section}}
\setcounter{section}{0}
\section{LLM-Driven Detail Synthesis}
\begin{figure}
\vspace{1em}
  \centering
  \centerline{\includegraphics[width=0.9\textwidth]{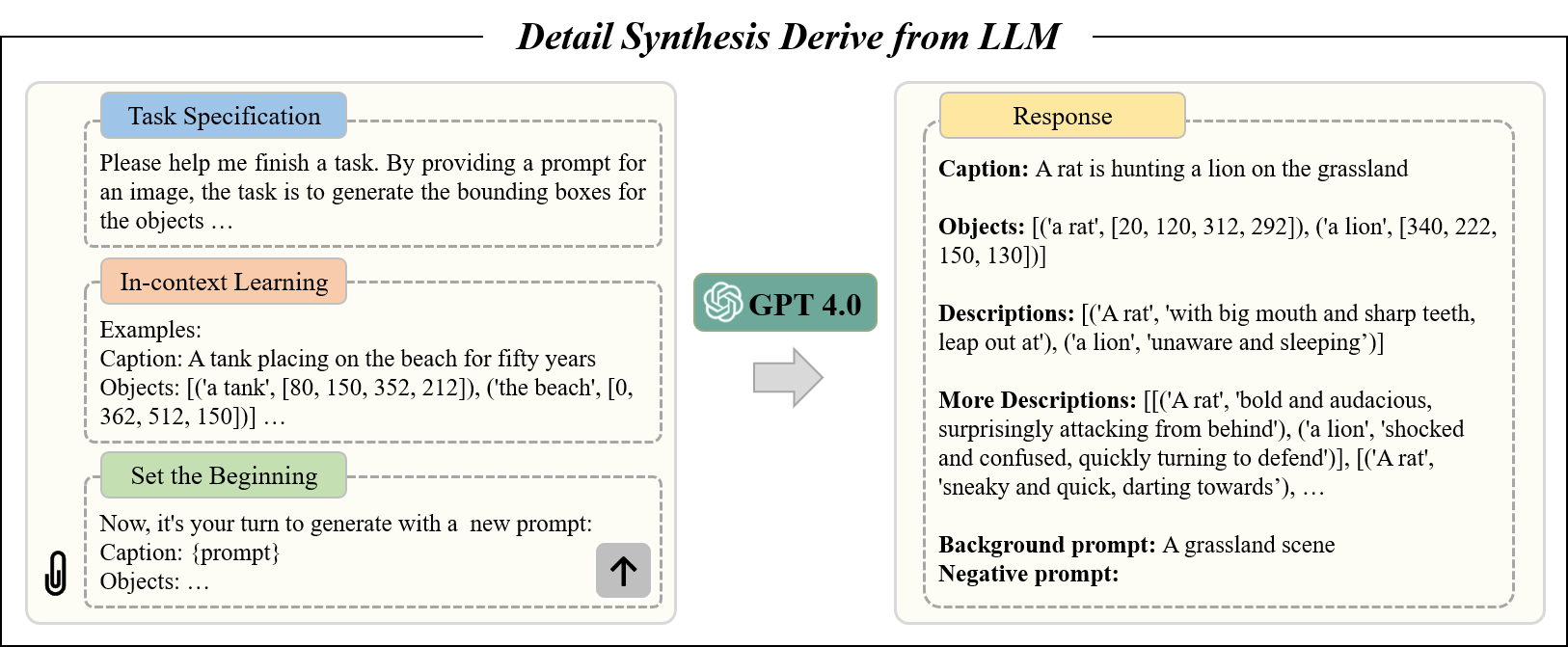}}
  \caption{\textbf{Detail Synthesis.} The illustration of the interaction with a LLM in our work.}
  \label{fig:part1}
\vspace{1em}
\end{figure}

In this work, as described in the Sec. \textcolor{red}{4.1} of the main paper, we emphasized that by leveraging LLMs, we have significantly enriched responses to encompass additional information, such as \textit{layout, detailed descriptions, background scenes,} and \textit{negative prompts}. To achieve this, we facilitated an interaction with a LLM as shown in \cref{fig:part1}. The input given to the LLM, depicted on the left side of the figure, includes detailed task specifications and in-context learning examples to enhance the LLM's comprehension. The response from the LLM, shown on the right, is rich with details extracted from the prompt. Notably, the descriptions are particularly crucial for our work, serving as indispensable information for the later image generation stage.

\vspace{-1em}

\begin{figure}
  \centering
  \centerline{\includegraphics[scale=0.42]{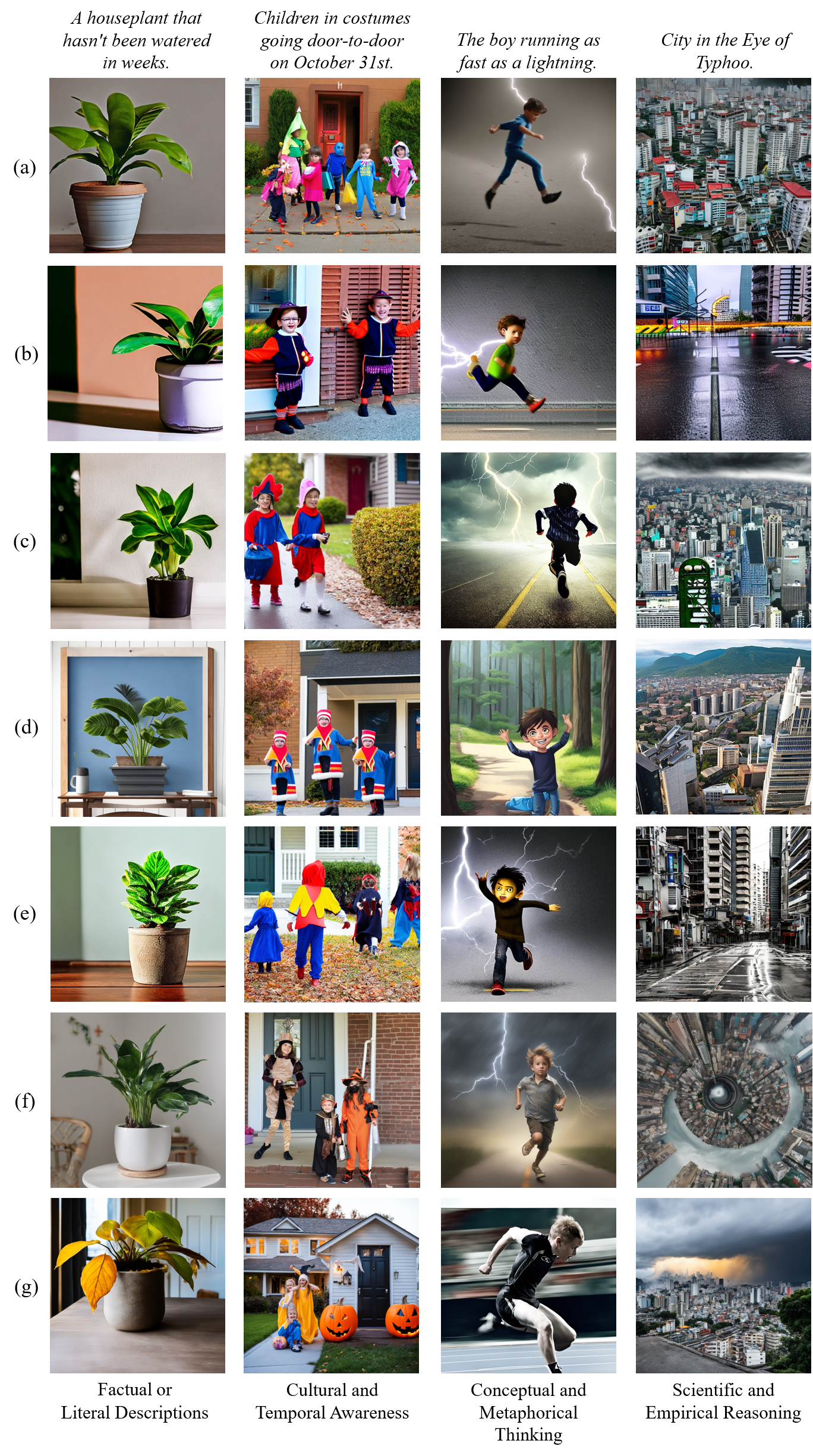}}
  \caption{More results on \textbf{\textit{Realistic and Analytical}.} The compared models include (a) Stable Diffusion, (b) MultiDiffusion, (c) AttendandExcite, (d) LMD, (e) BoxDiff, (f) SDXL, (g) Ours}
  \label{fig:sup_result1}
\end{figure}

\begin{figure}
  \centering
  \centerline{\includegraphics[width=0.9\textwidth]{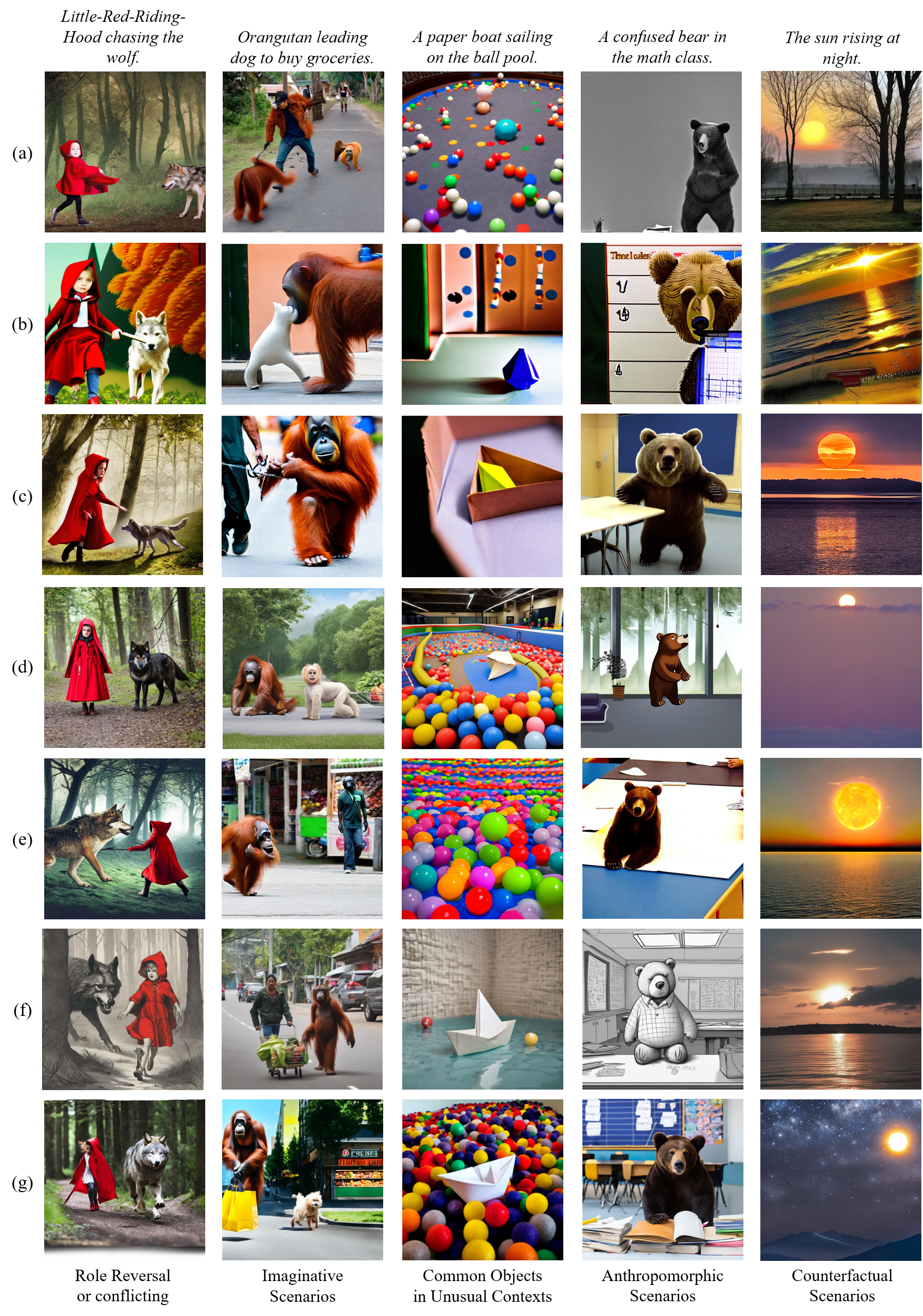}}
  \caption{More results on \textbf{\textit{Creativity and Imagination}.} The compared models include (a) Stable Diffusion, (b) MultiDiffusion, (c) AttendandExcite, (d) LMD, (e) BoxDiff, (f) SDXL, (g) Ours}
  \label{fig:sup_result2}
\end{figure}

\section{Qualitative Comparison on RFBench}
In \cref{fig:sup_result1} and \cref{fig:sup_result2}, we present additional qualitative examples to showcase the exceptional outcomes of our work. \cref{fig:sup_result1} shows the results under the category \textbf{\textit{Realistic and Analytical}}, while \cref{fig:sup_result2} shows the category \textbf{\textit{Creativity and Imagination}}. Both figures demonstrate that our method achieves more accurate editing results compared to other approaches.

\begin{figure*}[ht]
\vspace{-0.5cm}
  \centering
  \centerline{\includegraphics[width=0.8\textwidth]{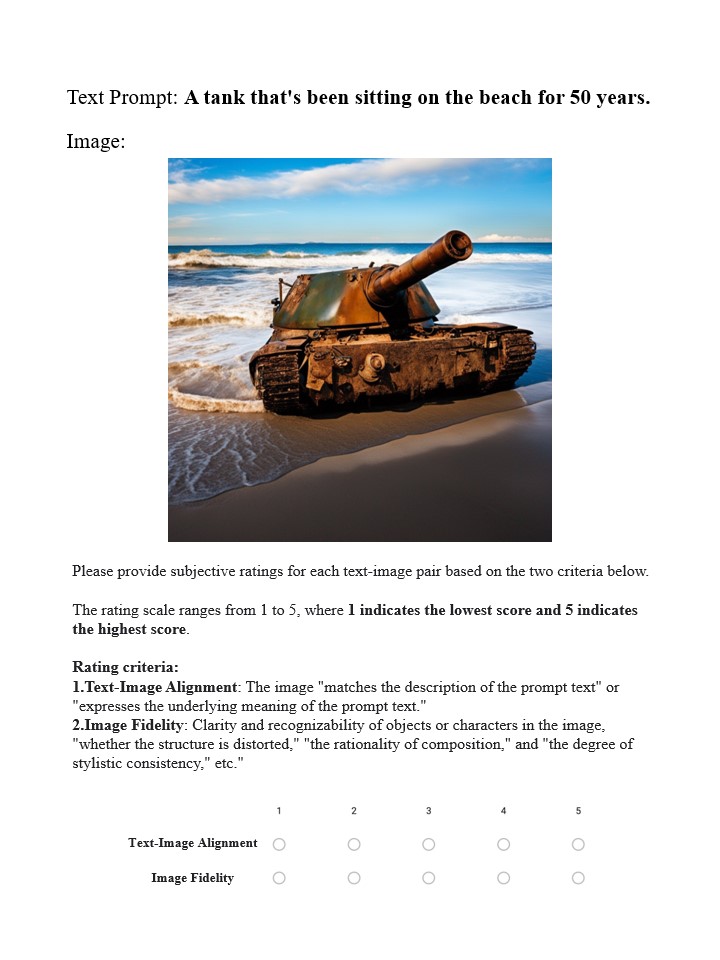}}
  \caption{Survey on Image-Text Alignment and Image Fidelity}
  \label{fig:survey_example}
\end{figure*}

\section{GPT4Score}

We follow the approach of T2I-Compbench, using Multimodal LLM (MLLM) to measure the similarity between generated images and input prompts. The key deviation lies in our observation that MiniGPT4, employed in T2I-Compbench, struggles to comprehend the surreal aspects of the images effectively. Therefore, we employ GPT4, a more powerful MLLM, as our new benchmarking model for evaluation, as mentioned in the Sec. \textcolor{red}{5.1} of the main paper.

Specifically, given a generated image and its prompt, we input both the image and prompt into GPT4. Subsequently, we pose two questions to the model: ``\emph{Describe the image}'' and ``\emph{Predict the image-text alignment score}'', the generated image is then assigned the final output score predicted by GPT4. For detailed prompts, please refer to the appendix of T2I-Compbench.

\section{Human Evaluation}
In the human evaluation process, as introduced in the Sec. \textcolor{red}{5.4} of the main paper, we request annotators to assess the correspondence between a produced image and the textual prompt employed to create the image. ~\cref{fig:survey_example} show the interfaces for human evaluation. The participants can choose a score from \{1, 2, 3, 4, 5\} and we normalize the scores by dividing them by 5. We then compute the average score across all images and all participants.

\captionsetup{skip=0.5\baselineskip}

\begin{table*}[!htbp] 
\setlength\tabcolsep{3pt} 
\centering
\tiny
\caption{The correlation between automatic evaluation metrics and human evaluation}
\resizebox{\textwidth}{!}{%
\begin{tabular}{>{\hspace{3pt}}ccccc<{\hspace{3pt}}}
\toprule
\multirow{2}{*}{Metrics} & \multicolumn{2}{c}{CLIPScore} & \multicolumn{2}{c}{GPT4Score} \\
\cmidrule{2-3} \cmidrule{4-5}
& $\tau$ ($\uparrow$) & $\rho$ ($\uparrow$) & $\tau$ ($\uparrow$) & $\rho$ ($\uparrow$) \\
\hline
\multicolumn{5}{c}{\textbf{Realistic} \& \textbf{Analytical}} \\
\hline
Scientific and Empirical Reasoning      & -0.4880 & -0.5946 & 0.6351 & 0.7157  \\
Cultural and Temporal Awareness         & -0.0476 & -0.1429 & 0.3273 & 0.3780  \\
Factual or Literal Descriptions         & 0.2333 & 0.3656 & 0.7620 & 0.8909  \\
Conceptual and Metaphorical Thinking    & -0.1952 & -0.1982 & 0.9234 & 0.9633  \\
\hline
\hline
\multicolumn{5}{c}{\textbf{Creativity} \& \textbf{Imagination}} \\
\hline
Common Objects in Unusual Contexts      & -0.2381 & -0.2857 & -0.5345 & -0.6124  \\
Imaginative Scenarios                   & 0.3752 & 0.6335 & 0.7265 & 0.8432  \\
Role Reversal or Conflicting            & 0.0476 & 0.1429 & 0.5040 & 0.5774  \\
Anthropomorphic Scenarios               & -0.1429 & -0.1429 & -0.5345 & -0.6124  \\
\bottomrule
\end{tabular}
}
\label{tab:table2}
\end{table*}

\section{Human Correlation of the Evaluation Metrics}
We adopt the methodology from T2I-Compbench, calculating Kendall's tau ($\tau$) and Spearman's rho ($\rho$) to evaluate the ranking correlation between CLIPScore, GPT4Score, and human evaluation. For better comparison, the scores predicted by each evaluation metric are normalized to a 0-1 scale. The human correlation results are presented in ~\cref{tab:table2}. These results indicate that CLIP underperforms in both categories, as discussed in Section \textcolor{red}{5.1} of the main paper. This underperformance may be due to CLIP's approach to image understanding, which is often too simplistic. Nevertheless, both metrics encounter challenges with \textbf{\textit{Creativity and Imagination}}, highlighting that although GPT4Score offers a broader understanding of images, accurately assessing creativity remains a difficult task for both.

\begin{figure}
\vspace{1em}
  \centering
  \centerline{\includegraphics[width=1\textwidth]{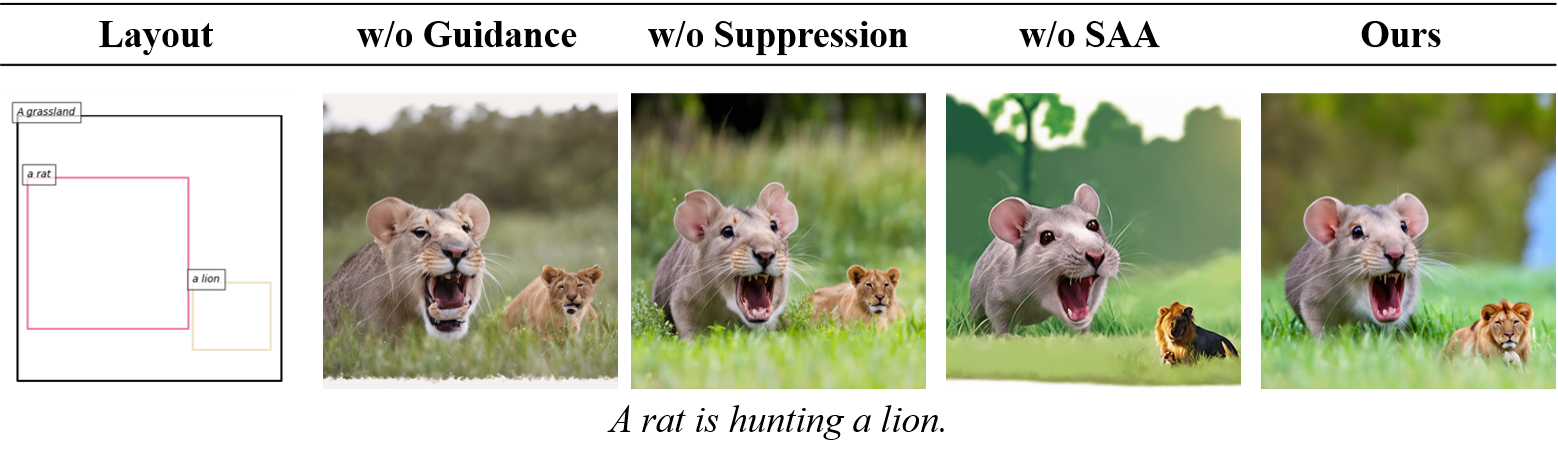}}
  \caption{Ablation study on various components in our work.}
  \label{fig:ablation_fig1}
\end{figure}

\section{Visualization of Ablation Study}
In addition to the quantitative results presented in our ablation study, we have also included visual examples to showcase the impact of different components in our work. As shown in \cref{fig:ablation_fig1}, the removal of guidance constraint and suppression constraint both causes the diffusion model to become muddled when dealing with multiple objects. Besides, eliminating the SAA module leads to unclear outcomes with the generated objects.

\subsection{Effect of the hyperparameter \(\beta\) of guidance constraint}
\begin{figure}
\vspace{1em}
  \centering
  \centerline{\includegraphics[width=1\textwidth]{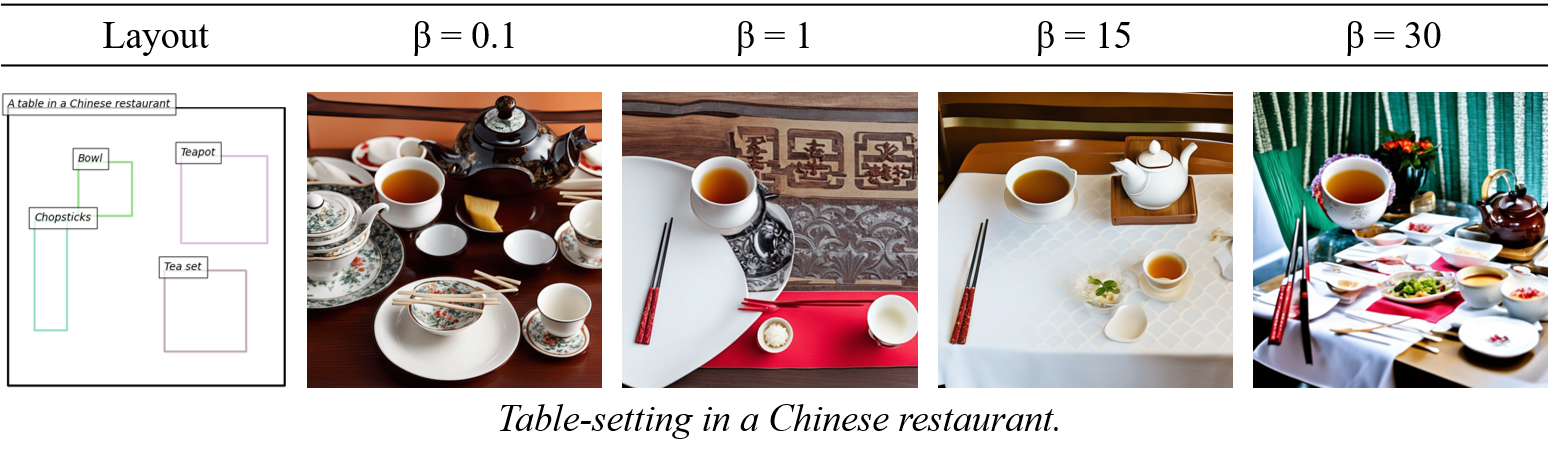}}
  \caption{Effect of the hyperparameter \(\beta\) of guidance constraint.}
  \label{fig:ablation_fig2}
\end{figure}

In our paper, we emphasize the critical role of the guidance constraint in integrating multiple objects into the background.  To underscore its significance, we performed an additional ablation study focusing on the hyperparameter \(\beta\), which influences the strength of guidance constraint. As shown in \cref{fig:ablation_fig2}, we varied \(\beta\) from 0.1 to 30 to observe the effects on the generated results. The findings reveal that an optimal \(\beta\) value (e.g., setting it to 15) ensures objects are accurately aligned with the layout and are of high quality. However, extreme \(\beta\) values, such as 0.1 or 30, disrupt the layout and diminish the overall quality of the generated images.

\end{document}